\documentclass{elsart}

\usepackage{natbib}

\usepackage{amsmath}
\usepackage{amssymb}
\usepackage{amsbsy}

\newcommand{\conju}{\mathrm{Conj}}

\newcommand{\PCR}{\mathrm{PCR5}}
\newcommand{\PCRmo}{\mathrm{PCR6}}

\begin{document}

\begin{frontmatter}



\title{Comments on ``A new combination of evidence based on compromise'' by K. Yamada}


\author{Jean Dezert$^{\mbox a}$, Arnaud Martin$^{\mbox b}$, Florentin Smarandache$^{\mbox c}$}

\address{$^{\mbox a}$ONERA, The French Aerospace Lab, 29 Av. Division Leclerc, 92320 Ch\^atillon, France}
\address{$^{\mbox b}$ ENSIETA, E$^3$I$^2$-EA3876, 2 rue Fran{\c c}ois Verny, 29806 Brest cedex 9, France}
\address{$^{\mbox c}$Chair of Mat. \& Sciences Dept., University of New Mexico, 200 College Road, Gallup, NM 87301, USA}

%
%
%
%

\end{frontmatter}


In a previous issue of this journal, \cite{Yamada08a} presents an interesting new model of combination of evidence called \emph{combination by compromise} based on previous works published in \cite{Yamada06a,Yamada06b}.

In the  theory of belief functions, one of the most important problems is the one of reassigning the conflicting belief mass, as  highlighted by the famous Zadeh's example, see \cite{Zadeh79}. To date, many combination rules have been developed, proposing a solution to this problem. \cite{Yamada08a} recalls some of them in his paper.

Basically, most combination rules are mainly based on a conjunctive operator and propose a specific way of redistributing the global (or the partial) conflicting belief mass among some elements of the power-set (or the hyper-power set) of the frame of discernment. Using the conjunctive rule means that if the experts agree (\emph{i.e.} their testimonies have a non-empty intersection), we consider them as reliable and if they are in conflict (empty intersection), at least one of the experts is considered unreliable  see \cite{Dubois88}. Then, the disjunctive combination rule can be employed instead see \cite{Dubois86}. However the disjunctive rule is generally not used because it deteriorates the specificity of the expert's responses, \emph{i.e.} the combined mass is usually less specific after the disjunctive fusion than the mass of each source taken separately. If the reliability of the experts is unknown, there are different techniques to estimate it \cite{Elouedi04a,Martin08a}.

The first idea  published in \cite{Yamada06a,Yamada06b} is based on the assumption that we do not have information on the reliability of the experts. So in order to define a compromise between the conjunctive and the disjunctive rules, Yamada proposes to transfer the basic belief assignments/masses $m_1(X).m_2(Y)$ given by two experts to $X\cap Y$, $X\cap Y^c$ and $X^c\cap Y$, even if $X$ and $Y$ do not conflict. The repartition is made proportionally to the basic belief assignment with a weight given by a ratio of cardinalities. In the more recent paper, \cite{Yamada08a} proposes a general form of repartition, with thresholds instead of the weights based on cardinalities.

After a review of criticisms against Dempster's rule and its advantages, \cite{Yamada08a} concludes on three possible ideas ``to combine two even hypotheses with the same reliability into one'' p~1698: 
\begin{enumerate}
\item Believe the common part of hypotheses (combination by exclusion, \textit{CBE}).
\item Believe the disjunction of hypotheses (``united part'' in \cite{Yamada08a}) (combination by union, \textit{CBU}).
\item Believe the common part strongly and the other part weakly (combination by compromise \textit{CBC}).
\end{enumerate}
The \textit{CBE} is similar to Dempster's rule. The CBU is similar to the disjunctive combination rule. \cite{Yamada08a} ``proposes the third approach, \textit{CBC} as a natural consensus. The basic idea is to share the mass $m_1(X)m_2(Y)$ among subsets included in $X\cup Y$.'' \cite{Yamada08a} examines ``three ways of sharing'':
\begin{enumerate}
\item The mass is shared between $X$ and $Y$.
\item The mass is shared between $X\cap Y$ and $X\cup Y$.
\item The mass is shared among $X\cap Y=C$, $X\cap Y^c=X_Y$ and $X^c\cap Y=Y_X$.
\end{enumerate}
For the first way of sharing, \cite{Yamada08a} proposes the equations (16) and (17) p~1698 given by:
$$m(X)=\frac{m_1(X)^2 m_2(Y)}{m_1(X) +  m_2(Y)} \qquad \text{and} \qquad m(Y)=\frac{m_1(X) m_2(Y)^2}{m_1(X) +  m_2(Y)}$$

For the second way of sharing, \cite{Yamada08a} proposes the equations (18) and (19) p~1699 given by: 
$$m(X\cap Y)=\delta m_1(X) m_2(Y)\qquad \text{and} \qquad m(X\cup Y)=(1-\delta) m_1(X) m_2(Y)$$
with $1\leq \delta \leq 1$. ``The value of $\delta$ could be chosen as the degree of overlapping between $X$ and $Y$, i.e. $\delta|X \cap Y| / |X\cup Y|$, where $|\bullet|$ means cardinality.''

For the third way of sharing, \cite{Yamada08a} proposes the equations (20), (21) and (22) p~1699 given by:
\begin{eqnarray*}
m(C)=\lambda_1 m_1(X) m_2(Y)\\
m(X_Y)=\lambda_2 m_1(X) m_2(Y)\\
m(Y_X)=\lambda_3 m_1(X) m_2(Y)
\end{eqnarray*}
where $0\leq \lambda_1, \lambda_2, \lambda_3 \leq 1$ and $\lambda_1+ \lambda_2+ \lambda_3=1$. The mass sharing of the \textit{CBC} is proposed in the section 4.3. in \cite{Yamada08a}. When $C=\emptyset$, the whole mass of $m_1(X) m_2(Y)$ is distributed to $X$ and $Y$ according to the first way of sharing. In \cite{Yamada08a}, the paper focuses on this third approach, but ``does not deny the qualification of the other subsets completely.''

However, we would like to recall some similarities between Yamada's two first ways of sharing the mass, and previously published combination rules not reported in the references of Yamada's  paper.

Whenever $X\cap Y=\emptyset$ and the mass that should be assigned to $X\cap Y$ is redistributed to $X$ and $Y$ proportionally to original masses $m_1(X)$ and $m_2(Y)$, Yamada's first way of sharing in \cite{Yamada08a} (equations (16) and (17) p~1698) is equivalent to the principle of sharing of the Proportional Conflict Redistribution rules \#5 (PCR5)  and \#6 (PCR6) for two experts\footnote{The PCR5 and PCR6 rules coincide for the two experts case.} published in \cite{Smarandache04c,Smarandache05a,Smarandache06b,Martin06b}. 

The  PCR5  and the PCR6 for two experts are given for two basic belief assignments $m_1$ and $m_2$ and for all $X \in 2^\Theta$, $X\neq \emptyset$ by:
\begin{eqnarray*}
\!\!\!\!\!m_\PCR(X)=m_\conju(X)~+ \!\!\!\!\!
\displaystyle
\sum_{\scriptstyle Y\in 2^\Theta, \,  X\cap Y = \emptyset}\left(\frac{m_1(X)^2 m_2(Y)}{m_1(X) \!+\!
  m_2(Y)}+\frac{m_2(X)^2 m_1(Y)}{m_2(X) \!+\! m_1(Y)}\!\right)\!\!,
\end{eqnarray*}
where $m_\conju(.)$ is the conjunctive rule. In the case of two experts, the Yamada's rule and the PCR5-PCR6 will be the same if all pairs of $X_i^1$ and $X_j^2$ chosen respectively from the expert 1 and 2 have the empty intersection ($X_i^1\cap X_j^2= \emptyset$, $\forall i, j$). For $M$ experts, if any pair of focal elements chosen from $M$ focal elements $X_i^1, \cdots, X_j^M$ have an empty intersection ($X_i^{k_1}\cap X_j^{k_2}=\emptyset$, \linebreak $\forall k_1,k_2=1,\cdots,M$, $k_1\neq k_2$ and $\forall i,j$) the Yamada's rule (equations (38), (39) and (40) p~1701 in \cite{Yamada08a}) is the PCR6 given explicitly for $X \in 2^\Theta$, $X\neq \emptyset$ by:
\begin{eqnarray*}
\!\!\!\!\!\!\!\!  \displaystyle m_\PCRmo(X)  =  \displaystyle m_\conju(X) + \sum_{i=1}^M
  m_i(X)^2 
  \!\!\!\!\!\!\!\!\!\!\!\!\!\!\!\!\!\!\!\! \displaystyle \sum_{\begin{array}{c}
      \scriptstyle {\displaystyle \mathop{\cap}_{k=1}^{M\!-\!1}} Y_{\sigma_i(k)} \cap X = \emptyset \\
      \scriptstyle (X_{\sigma_i(1)},...,X_{\sigma_i(M\!-\!1)})\in (2^\Theta)^{M\!-\!1}
  	\end{array}}
  \!\!\!\!\!\!\!\!\!\!\!\!
  \left(\!\!\frac{\displaystyle \prod_{j=1}^{M\!-\!1} m_{\sigma_i(j)}(X_{\sigma_i(j)})}
       {\displaystyle m_i(X) \!+\! \sum_{j=1}^{M\!-\!1} m_{\sigma_i(j)}(X_{\sigma_i(j)})}\!\!\right)\!\!,
\end{eqnarray*}
where $X_k \in 2^\Theta$ is the response of the expert $k$, $m_k(X_k)$ the associated belief function and $\sigma_i$ counts from 1 to $M$ avoiding $i$:
\begin{eqnarray*}
\label{sigma}
\left\{
\begin{array}{ll}
\sigma_i(j)=j &\mbox{if~} j<i,\\
\sigma_i(j)=j+1 &\mbox{if~} j\geq i,\\
\end{array}
\right.
\end{eqnarray*}
Indeed, in this special case, the equation (40) p~1701 in \cite{Yamada08a} is given by:
\begin{eqnarray*}
m_k(X_k)=\frac{m_k(X_k)}{\displaystyle \sum_{k=1}^{M} m_k(X_k)}\prod_{k=1}^{M} m_k(X_k).
\end{eqnarray*}

A presentation of PCR rules with many examples are proposed in \cite{Smarandache04c,Smarandache05a,Smarandache06b,Martin06b}. 

The second way of sharing proposed in \cite{Yamada08a} (equations (18) and (19)) is exactly the same as the mixed rule proposed in \cite{Martin07b}, equation (17), that also transfers the basic belief assignment $m_1(X)\cdot m_2(Y)$ even if $X$ and $Y$do not conflict. One of the proposed values of $\delta$ in \cite{Martin07b} is also given by $|X \cap Y|/|X \cup Y|$. In case of conflict $X\cap Y=\emptyset$, all the mass is transferred to $X \cup Y$. The mixed rule proposed in \cite{Martin07b} was inspired by \cite{Dubois88, Florea06a}. The recent paper \cite{Florea08a} presents more details on a robust combination rule.

To conclude, the proposed approach by Yamada 
can be an interesting alternative to the disjunctive rule and reconsiders the conjunctive rule. In very recent years many rules of combination have been proposed in the theory of belief functions. The choice of a rule is usually difficult and must be guided by the prior and the exogenous information (if available) related to the application one has to deal with.



%
%
%
%
%

\bibliographystyle{elsart-harv}
\bibliography{biblio}

\begin{thebibliography}{15}
\expandafter\ifx\csname natexlab\endcsname\relax\def\natexlab#1{#1}\fi
\expandafter\ifx\csname url\endcsname\relax
  \def\url#1{\texttt{#1}}\fi
\expandafter\ifx\csname urlprefix\endcsname\relax\def\urlprefix{URL }\fi

\bibitem[{Dubois and Prade(1986)}]{Dubois86}
Dubois, D., Prade, H., 1986. A set-theoretic view of belief functions - logical
  operations and approximation by fuzzy sets. International journal of General
  Systems 12~(3), 193--226.

\bibitem[{Dubois and Prade(1988)}]{Dubois88}
Dubois, D., Prade, H., 1988. Representation and combination of uncertainty with
  belief functions and possibility measures. Computational Intelligence 4,
  244--264.

\bibitem[{Elouedi et~al.(2004)Elouedi, Mellouli, and Smets}]{Elouedi04a}
Elouedi, Z., Mellouli, K., Smets, P., 2004. Assessing {S}ensor {R}eliability
  for {M}ultisensor {D}ata {F}usion {W}ithin {T}he {T}ransferable {B}elief
  {M}odel. IEEE Transactions on Systems, Man, and Cybernetics - Part B:
  Cybernetics 34~(1), 782--787.

\bibitem[{Florea et~al.(2006)Florea, Dezert, Valin, Smarandache, and
  Jousselme}]{Florea06a}
Florea, M., Dezert, J., Valin, P., Smarandache, F., Jousselme, A., March 2006.
  Adaptative combination rule and proportional conflict redistribution rule for
  information fusion. In: COGnitive systems with Interactive Sensors. Paris,
  France.

\bibitem[{Florea et~al.(2008)Florea, Jousselme, Bossé, and Grenier}]{Florea08a}
Florea, M., Jousselme, A., Bossé, E., Grenier, D., 2008. Robust combination
  rule for evidence theory. Information
  Fusion~(doi:10.1016/j.inffus.2008.08.007).

\bibitem[{Martin et~al.(2008)Martin, Jousselme, and Osswald}]{Martin08a}
Martin, A., Jousselme, A.-L., Osswald, C., July 2008. Conflict measure for the
  discounting operation on belief functions. In: International Conference on
  Information Fusion. Cologne, Germany.

\bibitem[{Martin and Osswald(2006)}]{Martin06b}
Martin, A., Osswald, C., 2006. A new generalization of the proportional
  conflict redistribution rule stable in terms of decision. In: Smarandache,
  F., Dezert, J. (Eds.), Applications and Advances of DSmT for Information
  Fusion. Vol.~2. American Research Press Rehoboth, Ch.~2, pp. 69--88.

\bibitem[{Martin and Osswald(2007)}]{Martin07b}
Martin, A., Osswald, C., July 2007. Toward a combination rule to deal with
  partial conflict and specificity in belief functions theory. In:
  International Conference on Information Fusion. Qu\'ebec, Canada.

\bibitem[{Smarandache and Dezert(2004)}]{Smarandache04c}
Smarandache, F., Dezert, J., December 2004. Proportional conflict
  redistribution rules for information fusion. Tech. rep.,
  http://xxx.lanl.gov/abs/cs/0408064v2.

\bibitem[{Smarandache and Dezert(2005)}]{Smarandache05a}
Smarandache, F., Dezert, J., June 2005. Information fusion based on new
  proportional conflict redistribution rules. In: International Conference on
  Information Fusion. Philadelphia, USA.

\bibitem[{Smarandache and Dezert(2006)}]{Smarandache06b}
Smarandache, F., Dezert, J., July 2006. Proportional conflict redistribution
  rules for information fusion. In: Smarandache, F., Dezert, J. (Eds.),
  Applications and Advances of DSmT for Information Fusion. Vol.~2. American
  Research Press Rehoboth, pp. 3--68.

\bibitem[{Yamada(2006{\natexlab{a}})}]{Yamada06b}
Yamada, K., 20-24 september 2006{\natexlab{a}}. On new combination of evidence
  by compromise. In: Joint 3rd International Conference on Soft Computing and
  Intelligent Systems and 7th International Symposium on advanced Intelligent
  Systems. Tokyo, Japan.

\bibitem[{Yamada(2006{\natexlab{b}})}]{Yamada06a}
Yamada, K., 6-8 september 2006{\natexlab{b}}. On new combination of evidence in
  evidence theory. In: 22nd Fuzzy System Symposium. Sapporo, Japan.

\bibitem[{Yamada(2008)}]{Yamada08a}
Yamada, K., July 2008. A new combination of evidence based on compromise. Fuzzy
  Sets and Systems 159~(13), 1689--1708.

\bibitem[{Zadeh(1979)}]{Zadeh79}
Zadeh, L., 1979. On the validity of {D}empster's rule of combination. Memo M
  79/24, Univ. of California, Berkeley, USA.

\end{thebibliography}

\end{document}